\documentclass[twocolumn]{IEEEtran}
\usepackage{soul}
\usepackage{lscape}
\usepackage{booktabs}
\usepackage{multirow}
\usepackage{authblk}
\usepackage{url}
\usepackage{silence}
\usepackage[font=scriptsize]{caption}

\usepackage[pdftex]{graphicx}
\usepackage{epstopdf}
\usepackage{pifont}
\usepackage[table,xcdraw]{xcolor}
\usepackage{array}
\usepackage{verse}
\usepackage{tikz}
 
\usepackage{longtable}
\usepackage{supertabular}
\usepackage{cite}
\usepackage{color}
\usepackage{amsmath}
\usepackage{lettrine}
\usepackage{hyperref}
\usepackage{multirow}
\usepackage{multicol}
\usepackage{dirtytalk}
\usepackage{color}

\usepackage[colorinlistoftodos,prependcaption,textsize=tiny]{todonotes}

\hypersetup{
 colorlinks=false,
 linkcolor=blue,
 filecolor=magenta,
 urlcolor=cyan,
}

\begin{document}

\title{Artificial Intelligence Based Prognostic Maintenance of Renewable Energy Systems: A Review of Techniques, Challenges, and Future Research Directions}

\author{
    \IEEEauthorblockN{Yasir Saleem Afridi \IEEEauthorrefmark{1}, Kashif Ahmad \IEEEauthorrefmark{2}, Laiq Hassan \IEEEauthorrefmark{1}}
    \\
    \IEEEauthorblockA{\IEEEauthorrefmark{1} Department of Computer Systems Engineering, University of Engineering and Technology, Peshawar, Pakistan.
    \\ {yasirsaleem,laiqhasan}@uetpeshawar.edu.pk} \\
    \IEEEauthorblockA{\IEEEauthorrefmark{2} Information and Computing Technologies (ICT) Division, College of Science and Engineering (CSE), Hamad Bin Khalifa University, Doha, Qatar
    \\ kahmad@hbku.edu.qa} \\
}
\maketitle

\begin{abstract}
\label{abstract}
Since the depletion of fossil fuels, the world has started to rely heavily on renewable sources of energy. With every passing year, our dependency on the renewable sources of energy is increasing exponentially. As a result, complex and hybrid generation systems are being designed and developed to meet the energy demands and ensure energy security in a country. The continual improvement in the technology and an effort towards the provision of uninterrupted power to the end-users is strongly dependent on an effective and fault resilient Operation \& Maintenance (O\&M) system. Ingenious algorithms and techniques are hence been introduced aiming to minimize equipment and plant downtime. Efforts are being made to develop robust Prognostic Maintenance systems that can identify the faults before they occur. To this aim, complex Data Analytics and Machine Learning (ML) techniques are being used to increase the overall efficiency of these prognostic maintenance systems.

This paper provides an overview of the predictive/prognostic maintenance frameworks reported in the literature. We pay a particular focus to the approaches, challenges including data-related issues, such as the availability and quality of the data and data auditing, feature engineering, interpretability, and security issues. Being a key aspect of ML-based solutions, we also discuss some of the commonly used publicly available datasets in the domain. The paper also identifies key future research directions. We believe such detailed analysis will provide a baseline for future research in the domain. 
\end{abstract}

\section{Introduction}
\label{sec:introduction}
The set of various activities performed on a system to smoothly execute its operational state is known as System Health Monitoring (SHM) or Condition Based Monitoring (CBM). CBM is limited to only observing the current operational states of the system. Repair and maintenance actions are hence generated, solely based on these current states. When these current operational states are augmented with the prediction of future failure states, the process is then termed as Predictive Maintenance or Prognostics \cite{martin1994review}. Prognostics is a very complicated procedure and needs precise, adaptive, and intuitive models. These models are utilized to predict the future failure states of the system. Hence, prognostics help to reduce the machine downtime and increase the Remaining Useful Life (RUL) of the equipment. 

\begin{figure}[!ht]
\graphicspath{ {./Figures/} }
    \centering
	\includegraphics[width=0.45\textwidth]{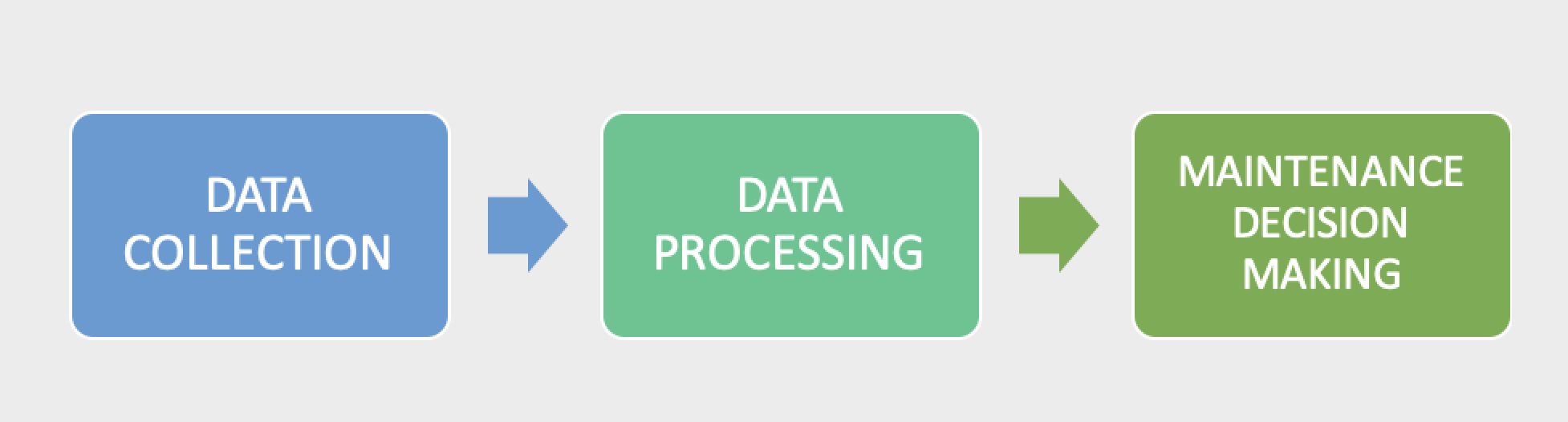}
	\caption{The three steps of a CBM process.}
	\label{fig:CBM_process}
\end{figure}

The conventional maintenance techniques mostly rely on a reactive approach or unplanned maintenance i.e. the maintenance activity is conducted once the break down had occurred. This is also known as Run-to-Failure Maintenance. Another approach, known as Preventive Maintenance (PM), where a machine is inspected at predefined time intervals. PM is purely periodic and does not consider the actual health of a machine. Instead of the prevailing rapid technological development, systems and machines are becoming more and more complex. This complexity has given rise to the need for enhanced reliability and lesser downtimes, thereby exponentially increasing the costs associated with PM, hence, making it a less feasible option. The complexity and risks involved in the maintenance of power systems demand a more efficient, reliable, and robust maintenance system, such as the CBM \cite{martin1994review}. CBM is a maintenance technique that tends to reduce the frequency of the unscheduled corrective or scheduled PM by actively monitoring the machine's current operating states and highlighting abnormal behaviors of the system. Hence, CBM not only lowers maintenance costs but also helps to reduce plant or machinery downtime. A CBM process mainly resonates around three steps \cite{lee2004integrated} as depicted in Figure \ref{fig:CBM_process}.

The data collection step involves monitoring the current operational states of the system and gathering the relevant data. Secondly, the collected data is pre-processed by normalizing and removing outliers. The pre-processed data is easy to understand and it also facilitates the decision-making process. Lastly, Machine Learning (ML) algorithms are then applied to the pre-processed data to decide to carry out a maintenance activity \cite{lee2004integrated}.
We noted that a CBM is called a diagnostic system when it only incorporates conventional reactive means of maintenance, whereas it is called a prognostics system if it proactively identifies the faults before their occurrence. The prognostics system predicts the estimated amount of time before an impending fault may occur. Hence, prognostics is a prior event analysis whereas diagnostics is a posterior event analysis. In the power generating systems, where reducing machine downtime is inevitable, prognostics outperforms the diagnostics systems. However, when the prognostic system fails to predict a fault, diagnostics are required to isolate the system fault and rectify it. A CBM approach hence incorporates either one or both the diagnostic and prognostic techniques for fault identification.

In the modern world, rather than a single power generation source, countries are now relying on multiple power-generating sources that can work in coordination and integrate seamlessly with the deployed transmission and distribution network. Research is being conducted to develop new equipment that can cope up with the conundrum of multiple generation sources and work effectively with the smart grids, being deployed globally. Being a key aspect of smart grids, predictive maintenance has been a keen area of interest for researchers. Several interesting solutions covering different aspects of predictive maintenance have been proposed in the literature. In this paper, we provide a detailed overview of the existing literature on predictive maintenance in three different sources of renewable energy including hydroelectric, wind, and solar with a particular focus on techniques, challenges, and future research directions.



\subsection{Scope of the Survey}
\begin{figure}[!ht]
\graphicspath{ {./Figures/} }
    \centering
	\includegraphics[width=0.45\textwidth]{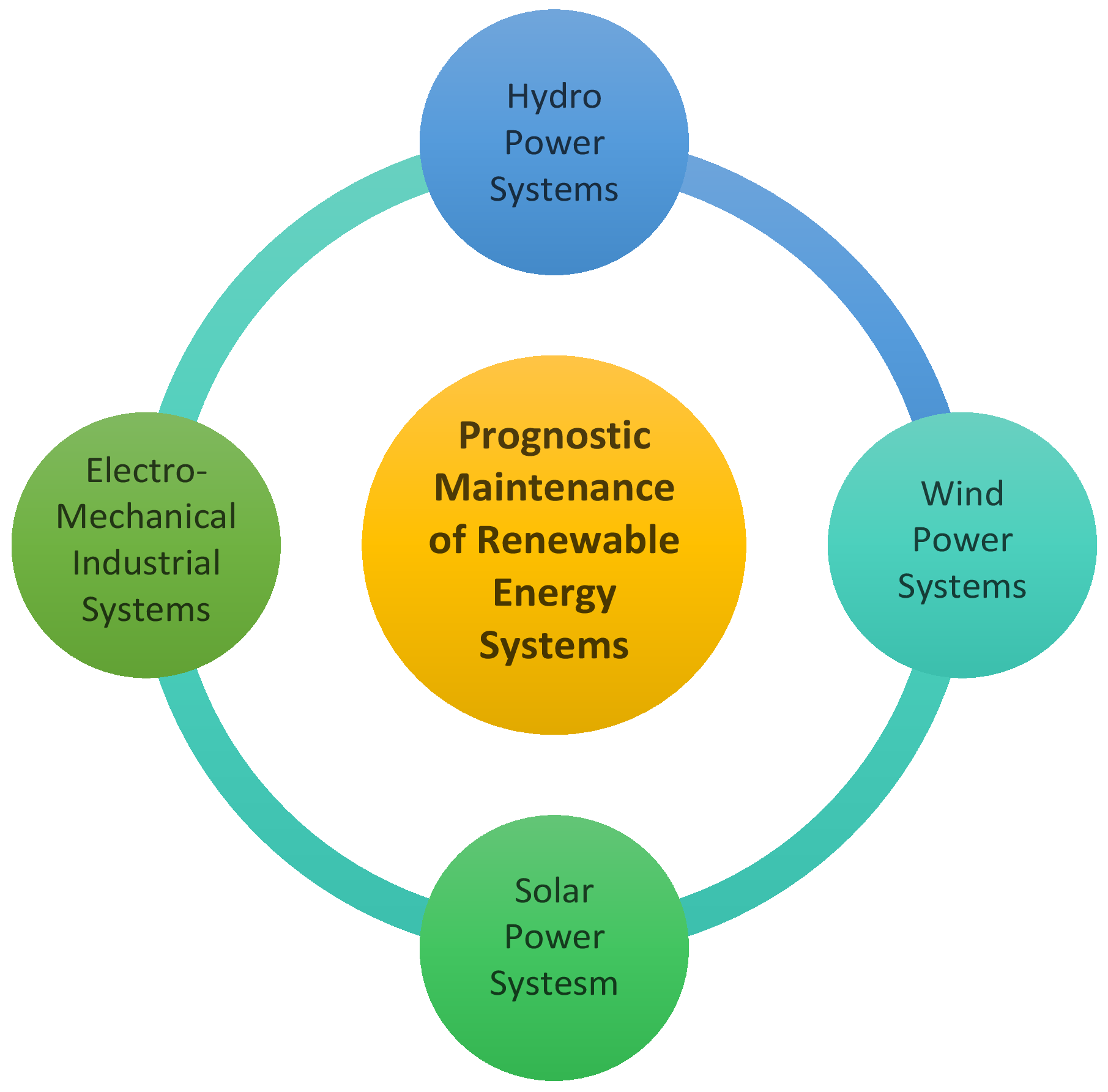}
	\caption{The scope of this survey.}
	\label{fig:Scope}
\end{figure}

Since the applications of CBM and Prognostic Health Management (PHM) are very diversified, this paper, therefore, encompasses the CBM and PHM techniques and procedures adopted by various researchers in renewable energy (wind, hydro and solar power) systems and industrial electro-mechanical systems. The paper not only explores the methodologies adopted during the early researches in the area of PHM but also analyzes more recent challenges and trends in this domain. The paper also identifies the current limitations and challenges and gives directions for future research to fill in the literature gaps. Figure \ref{fig:Scope}  visually depicts the scope of the paper.

\subsection{Related Surveys}
Pursuant to the prevailing advancements in the renewable energy sector at a very fast pace, complex systems are being designed, which require more effective and efficient Operations and Maintenance (O\&M) schemes. With a globally increased demand and dependency on energy, plant downtimes and outages are highly unaffordable. Hence, the need for robust and proactive O\&M systems has increased like never before. Therefore, the development of effective prognostic maintenance systems is always in the spotlight. Extensive research is being conducted in this area. There are also some existing surveys covering different aspects of the domain. The surveys and reviews being conducted in this domain are either generically relevant to CBM processes in multiple domains and industries \cite{article} or specifically related to a single domain like wind power turbines \cite{maldonado2020using}. For instance, Quatrini et. al. \cite{machines8020031} conducted a very extensive survey pertinent to the research conducted on CBM processes. The survey focuses on four main research areas including implementation strategies, operational aspects, inspections, replacements, and prognosis.
Jorge Maldonado et. al. \cite{maldonado2020using} presented a systematic literature review on the use of Supervisory Control and Data Acquisition (SCADA) data and Artificial Intelligence (AI) techniques for carrying the CBM of wind turbines. A detailed survey on the prognostics and health management of machine tools with the perspective and analysis of the Industry 4.0 maintenance policy has been conducted by Marco Baur et. al. \cite{baur2020review}. Shaomin Wu et. al. \cite{wu2020machine} has categorized the publications having the top 5\% citations and employ classical ML methods, such as Support Vector Machines (SVMs), random forests, and cluster analysis, for the reliability analysis and optimization of maintenance procedures. Cherry Bhargava et. al. \cite{bhargava2020review} have carried out a critical analysis of various researches on the statistical, empirical, and intelligent tools and techniques used for the monitoring of electronic components in green manufacturing. A comprehensive survey of the literature focused on the signals and signal processing methods for conducting the CBM and fault diagnostics of wind turbines has been carried out by Wei Qiao et. al. \cite{qiao2015survey}. Similarly, a general overview of the techniques and procedures used in the condition monitoring of wind turbines has been provided in \cite{tchakoua2014wind}. The authors have also outlined a relationship between concepts and theories and analyzed the state-of-the-art trends and future challenges in wind turbine condition monitoring. Likewise, Mengyan \cite{NIE2013287} identified the challenges and opportunities in the condition monitoring and prognosis of wind turbine gearboxes.  They also discussed the importance and role of different sensors in wind turbine gearbox condition monitoring, along with their fault diagnosis. On the other hand Yuri Merizalde et. al. \cite{merizalde2019maintenance} identified and classified different mathematical models used for decision making at strategic, tactical, and operational levels of wind turbine maintenance.

It is evident from the literature that most of the surveys and reviews conducted in the area of CBM and prognostics are either too specific, focusing on a single domain, or too generic, focusing on multiple industries. Furthermore, surveys pertinent to the CBM processes in hydropower and solar power systems are very rare to find.  To this aim, this paper provides an in-depth and detailed survey of the literature on CBM and prognostics of renewable energy systems, thereby encompassing the Hydro Power, Wind Power, and Solar Power systems with a particular focus on approaches, challenges, such as security, interpretability, and data-related issues, and future research directions. Table \ref{tab:comparison_other_surveys} provides an overview of some of the existing related surveys.

\begin{table*}[]
\centering
\caption{A summary and comparisons against existing surveys on the topic in terms of year, domain, and main focus of the papers.}
\label{tab:comparison_other_surveys}
\scalebox{0.85}{
\begin{tabular}{|p{.2cm}|p{.5 cm}|p{4cm}|p{11cm}|}
\hline
\multicolumn{1}{|c|}{Refs.} & \multicolumn{1}{c|}{Year} 
&\multicolumn{1}{c|}{Domain} &\multicolumn{1}{c|}{Main Focus} \\ \hline

\cite{machines8020031}& 2020 & Multiple Industries& This survey has focused four main areas in the research conducted on CBM processes including (i) Implementation Strategies, (ii) Operational Aspects, (iii) Inspections and Replacements, and (iv) Prognosis.\\ \hline
\cite{article}& 2016&Multiple Industrial Systems &Provided a preliminary study of maintenance optimization applications implemented in different industries and guidelines for developing an implementation plan for CBM programs. \\ \hline
\cite{baur2020review}& 2020& Machine Tools&Provided a detailed survey on the prognostics and health management of machine tools with the perspective to the Industry 4.0 maintenance policy. \\ \hline
\cite{wu2020machine}&2020 & Reliability and Maintenance Engineering& Categorized the recent research papers that have used Support Vector methods, Random Forests and Cluster Analysis for the reliability analysis and optimization of maintenance procedures.\\ \hline
\cite{bhargava2020review}&2020 & Electronic Components& Carried out a critical analysis of various researches on the statistical, empirical and intelligent tools and techniques used for the condition based monitoring of electronic components in green manufacturing.\\ \hline
\cite{maldonado2020using}&2020 & Wind turbines& Focuses on the use of SCADA data and AI for carrying out the CBM of wind turbines.\\ \hline
\cite{merizalde2019maintenance}&2015 & Wind turbines& Identified and classified different mathematical models used for decision making at strategic, tactical and operational levels of wind turbine maintenance.\\ \hline
\cite{qiao2015survey}& 2015& Wind turbines& Conducted a survey of the researches focused on the signals and signal processing methods for carrying out the CBM and fault diagnosis of wind turbines.\\ \hline
\cite{tchakoua2014wind}&2014 & Wind turbines& Provided a general review of the techniques and procedures used in the condition monitoring of wind turbines with a focus on trends and future challenges.\\ \hline
\cite{NIE2013287}& 2013& Wind Turbines& Identified the challenges and opportunities in the condition monitoring and prognosis of wind turbine gearboxes, with a particular emphasis on the application of sensor data in wind turbine gearbox condition monitoring.\\ \hline
\textbf{This Work}& 2021& Electromechanical industrial systems, Hydro, Wind, and Solar Power Systems& Provides an extensive and detailed survey of the early researches conducted as well as state-of-the-art techniques and procedures adopted in the area of CBM and Prognostic Maintenance with an emphasis on renewable energy including hydro power, wind power and solar power systems. \\ \hline

\end{tabular}
}
\end{table*}

\subsection{Contributions}
This paper provides a detailed survey of the challenges faced in the prognostics and CBM of electro-mechanical systems and the approaches, methodology, and techniques adopted to address these challenges. Instead of relying on a single domain, the paper provides a diversified and extensive review of the literature pertinent to CBM and PHM of various types of renewable energy systems including hydro, wind, and solar power. Along with the performance evaluation of various ML algorithms used in fault prognostics, this paper also discussed the shortcomings and disadvantages of the adopted techniques and provides future research directions. We pay particular attention to the challenges faced in successful deployment of predictive maintenance algorithms, including data-related issues, such as data availability, quality of data and data auditing, feature engineering and selection, interpretability, security, and adversarial attacks. 
A summary of the main contributions of the paper is as follows:

\begin{itemize}
 \item We provide an in-depth analysis of the literature on various techniques and procedures for carrying out PHM of industrial and renewable energy systems.
 \item Along with the early researches in the domain, this paper also analyzes the recent literature on major challenges in carrying out the predictive maintenance of renewable energy systems including wind, hydro and solar power.
 \item The paper also provides a taxonomy of various ML techniques used for carrying out PHM.
 \item Benchmark datasets that can be used for training and testing of the model are also being identified in this paper.
 \item Moreover, we have also identified the shortcomings, challenges, open issues, and provided the future research directions.
\end{itemize}

The rest of the paper is organized as follows: Section \ref{sec:Legacy_of_CBM} provides a detailed discussion of early researches conducted pertinent to the prognostic maintenance techniques of various electro-mechanical systems. Section \ref{sec:HM} is focused on more recent research relevant to the prognostic maintenance of renewable energy systems including Hydro, Wind, and Solar Power Projects. Section \ref{sec:taxonomy_ML} provides an overview of some commonly used ML techniques for predictive maintenance. Section \ref{sec:datasets} describes some of the publicly available benchmark datasets for training and evaluation of predictive maintenance frameworks. Section \ref{sec:discussion} discusses the key lessons learned from the literature. Finally, Section \ref{conclusion} concludes the paper.

\begin{figure*}[!h]
\centering
\graphicspath{ {./Figures/} }
\includegraphics[width=.99\linewidth]{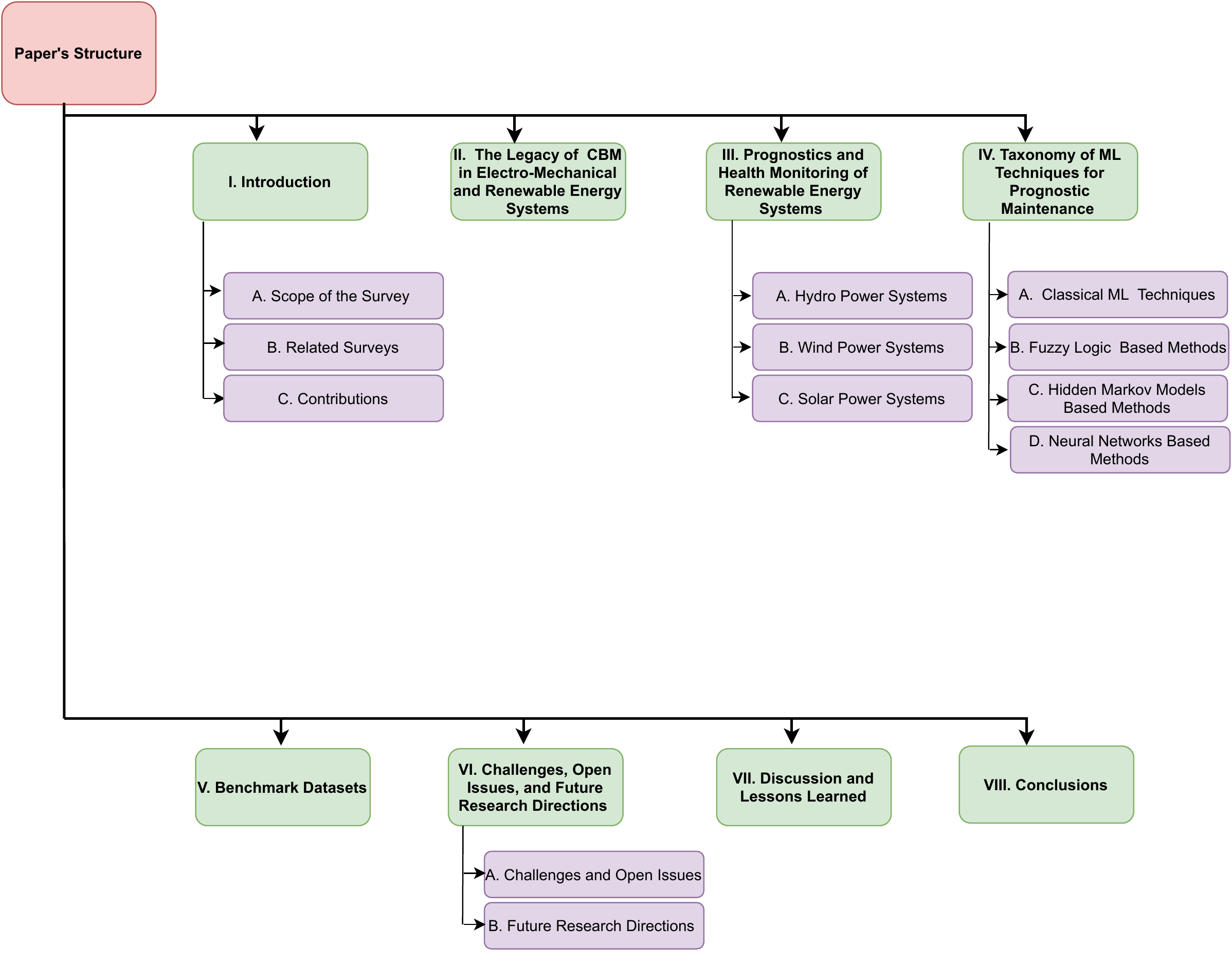}
\caption{Structure of the survey.}
	\label{fig:structure}
\end{figure*}
\section{The Legacy of CBM in Electro-Mechanical and Renewable Energy Systems}
\label{sec:Legacy_of_CBM} 
CBM, SHM, and their integrated functions have been in the spotlight of research and are being implemented for quite some years now. With each passing day, these techniques have evolved significantly. Their governing philosophy, implementation strategies, technological advancements, modeling techniques, algorithms, and the continually changing necessities have been an integral part of the said evolution process. As explained by Kinclaid et al. \cite{kinclaid1987evolution}, this evolution process pertinent to the SHM follows a certain perspective in a very interesting chronological fashion. 

As part of the initial efforts in the domain, Wu et al. \cite{wu2007neural} proposed a probabilistic replacement policy aiming the minimization of average replacement cost over the operational life of a component/machine. To this aim, an Artificial Neural Network (ANN) based framework was developed and trained using real-time sensory vibrations data to predict the remaining life percentage of a machine. Optimal replacement strategies were being proposed based on a cost matrix model. However, in this research, the distribution of marginal life and cost matrix was limited to intervals of ten percentile rather than being continuous. Tan et al. \cite{tan2008framework}, on the other hand, proposed a framework for predictive maintenance modeling for multi-state systems. They proposed a framework for scheduling the predictive maintenance of multi-state systems. An estimation of the failure times through performance degradation trends was used to derive maintenance schedules from the system perspective. A statistical model for the estimation of the predictive maintenance was designed using HMMs. The model simulation results depicted a promising impact of the maintenance quality and maintenance schedules on the system reliability and mean performance characteristics. Hence, it was found that the system replacement time was extended manifold by a very slight improvement in the quality of the maintenance.  However, unlike most real-life systems, the proposed model had a single mode of failure and it considered all the elements to be independent. The model, therefore, could be extended to consider the effect of dependent multi-modal failures.

Kalgren et al. \cite{kalgren2007application} presented a technical approach for prognostic health management of digital electronic systems. The integrated collaborative diagnostic and prognostic techniques from various disciplines of engineering including modeling of damage accumulation, physics of failure, reliability statistics, signal processing, feature extraction, and automated reasoning algorithms. These advanced amalgamated algorithms achieved the most profitable decisions pertinent to the overall health of the electronic systems by using intelligent data fusion architectures to optimally combine sensor data with probabilistic component models. Carrying an in-depth understanding of semiconductor devices and being one of the pioneer researches, it paved the way for the development of prognostic HMS for a wide range of digital electronic systems.  Likewise, Pecht et al. \cite{pecht2007prognostics} integrated various effective prognostic health management techniques and proposed an optimal and efficient system for calculating the Remaining Useful Life (RUL) of electronic devices. The study introduced a methodology for monitoring the life consumption process and assessed the life already consumed using physics-based stress and damage models. Maintenance schedules were generated based on the RUL of the component calculated through forecasting models and regression analysis.

Lu et al. \cite{lu2007predictive} devised a predictive condition-based maintenance for continuously deteriorating systems. The authors considered the system to be continuously degrading as a stochastic gradient process. The system deteriorations were recursively modeled and forecasted using the structural time series along with the state-space modeling and Kalman filtering methods. The predicted deterioration state of the system and a pre-defined threshold of failure were used to calculate the failure probabilities. Maintenance decisions were hence made based on the predicted failure probabilities, cost of maintenance, and financial implications due to the performance deterioration of the system. Similarly, Zhou et al. \cite{zhou2007reliability} developed a reliability-centered predictive maintenance scheduling scheme for a continuously monitored system subject to degradation. In this work, the reliability value when exceeded a pre-defined threshold $R$, a call for preventive maintenance was generated. During various maintenance cycles, the reliability value of the system was predicted by deriving the hazard rate of the system directly from condition-based preventive maintenance (CBPM). The value of the threshold $R$ was optimized to reduce the maintenance cost per unit time in terms of RUL of the machine.

Zhaohui et al. \cite{li2002integrated} proposed certain integrated maintenance features of hydro turbine governors intending to be implemented at the three Gorges Hydro Power Project, in China. The researchers studied and analyzed the maintenance features of the governor and proposed an expert system that functioned in parallel to the governor’s tasks. These features made promising contributions towards the increased availability of the Hydro Turbine governing system.  Li et al. \cite{li2007optimal} designed a Hydro Power Optimal Maintenance Information System (HOMIS). HOMIS was implemented at Gezhouba Hydro Power Plant, China. The optimal Condition Monitoring system was developed by integrating the Supervisory Control and Data Acquisition (SCADA) and the Management Information System (MIS) modules of the Power Plant. 


Engin \cite{engin2007prediction} predicted the degradation in the relative efficiency of centrifugal pumps while handling slurries by establishing a correlation with the data generated from tests conducted on a specially designed test facility. The applicability of neural networks for the prediction of the same task was then examined. The author also conducted a comparative evaluation of the predictions made both by using the correlation method and ANNs. Similarly, Su et al. \cite{su2007induction} developed a Neural Network based condition monitoring of induction machines. The output of the induction motor system in the vibration spectra was modeled using a multilayer perceptron (MLP). Training of the network was done using the Lavenberg-Marquardt (LM) algorithm based on the Gauss-Newton method. The experimental results generated by the system were very promising. The features of the signals received were more distinct as compared to the conventional methods. The proposed methodology can also be used for condition monitoring of the motors in real-time through online frequency analysis. Likewise, Sugumaran et al. \cite{sugumaran2007feature} identified optimal features for classification purposes from the data generated by a 3000Hp DC motor. Based on the features selected, Proximal SVMs (PSVMs) was then used as a classifier to efficiently identify the faults in the roller bearings of the motor. PSVMs has the advantage of faster learning over the conventional SVMs and hence works more effectively in fault diagnostics. However, the algorithm requires a large number of data points.
Widodo et al. \cite{widodo2007combination} used a combination of Independent Component (ICA) and SVMs for intelligent faults diagnosis of induction motors. ICA was used both to identify the features and to carry out the dimensionality reduction. The training of the SVMs was conducted using the Sequential Minimal Optimization (SMO) algorithm. Furthermore, keeping in view the diversified types of faults in an induction motor, a multi-class SVMs-based classification was used. The combination of ICA and SVMs generated very promising results in fault identification.

Wilkinson et al. \cite{wilkinson2007condition} evaluated the condition monitoring approaches for the off-shore Wind Turbines to detect faults. A 30KW test rig was constructed imitating the exact features that of a wind turbine drive train. Various condition monitoring approaches were investigated and it was found that by using the appropriate signal processing techniques, properties of the gearbox, and load variations,  faults in the coil could be detected. Similarly, Wiggelinkhuizen et al. \cite{wiggelinkhuizen2007condition} investigated the effectiveness of the already developed Condition Monitoring techniques against Offshore Wind Farms. Analysis of the data gathered from the wind turbine through Fast Fourier Transform (FFT), Root Mean Square (RMS), and Wavelet Transformation was carried out to find out the turbine degradation. Deviations from the normal behavior were detected by analyzing the de-trending of SCADA data. Time series analysis and vibration measurements successfully indicated shaft misalignments. It was being concluded that for all types of measurements, since a large amount of data is being generated by a wind farm, it is difficult for the operational staff to interpret it. Hence, a dedicated team of data analysts is required to carry out the proposed data analysis.

Yuan et al. \cite{yuan2007fault} proposed the use of Artificial Immunisation Algorithm for optimization of parameters in SVMs based fault diagnosis, that on the other hand is performed with human intervention. The elimination of human interference for parameters optimization resulted in effectively overcoming the problems pertinent to premature convergence and local optimum. The Artificial Immunisation Algorithm – Support Vector Machine (AIA-SVMs) was found to be very efficient in diagnosing the multi-class faults. AIA-SVMs had a higher recognition rate than the conventional SVMs algorithm.

Hu et al. \cite{hu2007fault} developed a novel method for the diagnosis of faults in rotating machinery. The faults based on the vibration signals data were being diagnosed by using an Improved Wavelet Package Transform (IWPT), a distance evaluation technique, and SVMs. The experimental results depicted that the novel SVMs ensemble had a much higher average fault diagnosis accuracy than the conventional SVMs.

Xu et al. \cite{xu2007gas} integrated the wavelet transform with the neural networks for fault diagnosis in the Gas Turbines. This integration helped to overcome the limitations that the conventional backpropagation neural network holds. The experimental test results depicted that the wavelet neural network had improved diagnostic accuracy and learning speed than the conventional backpropagated neural networks. Similarly, Qingyang et al. \cite{qingyang2008gas} used an Adaptive Resonance Theory (ART2) based neural network for fault diagnosis in Gas turbines. The shock sensor installed on the gas turbine generated a shock signal. The received shock signal was then disintegrated into different frequency bands using the spectrum analysis and discrete wavelet transform. Once the signal was decomposed, the spectrum energy for each signal was analyzed and the associated structure of each spectrum was extracted. Samples, that best reflected the faults were then selected. The samples were then inputted into the network for effective classification and diagnosis of the faults. ART2 Neural Networks proved to be very adaptive to non-stable environments and showed a rapid learning rate, despite being purely unsupervised.

Shalev et al. \cite{shalev2007condition} developed a new methodology for improved Fault Tree Analysis (FTA) and reliability and safety calculations; the Condition-Based Fault Tree Analysis (CBFTA). In the CBFTA approach, reliability calculations were made a part of the condition monitoring thereby enhancing the operational safety of the industrial systems. In CBFTA, the condition-based failure rate was calculated after every periodic maintenance cycle, unlike the FTA, where the failure rate is considered constant throughout the remaining useful life of the machinery. Using the condition-based failure rate and the FTA, updated system failure rate and system reliability value were obtained. Periodic recalculation of these values resulted in the desired CBFTA. 

This section revealed an in-depth analysis of all the research conducted during the previous decade in the field of CBM for various electro-mechanical systems including induction motors, wind, gas, and hydro turbines and industrial systems. Various statistical and ML tools have been used to carry out the diagnostic process. More advanced techniques, including the wavelet, transforms in combination with the Artificial Neural Networks have also been in practice. SVMs and HMM have also been adopted for effective fault prediction. Thus, various approaches including state-of-the-art techniques have been proposed for system fault diagnosis based on the technological complexity of the machines. However, it has been observed that in most scenarios limited data has been considered due to the computational expensiveness of the algorithms. Consequently, it reduced the efficiency of the proposed algorithms for carrying out the prognostics. Furthermore, in some of the cases, real-life implementation of methodologies to gather data was highly expensive, thus not feasible.

\section{Prognostics and Health Monitoring of Renewable Energy Systems}
\label{sec:HM}
This section is focused on more recent studies conducted in the area of CBM and SHM pertinent to renewable energy systems. A detailed analysis of various proposed methodologies for the prognostics and health monitoring of hydro, wind, and solar power projects is provided in the next sub-sections.


\subsection{Hydro Power Systems}
Hydropower has been used as one of the first renewable sources of energy. In almost every country, hydropower projects have always been relied upon to meet the baseload demands of the grid. Plant availability and reduced downtimes are inevitable when it comes to Hydro Power Projects (HPPs). To ensure that, efficient Operation and Maintenance (O\&M) systems need to be deployed. With the current pace of technological advancement resulting in more complex equipment, a global paradigm shift from conventional preventive maintenance systems to more advanced diagnostic and prognostic maintenance systems is required. To this aim, several efforts have been made \cite{kande2017rotating,wang2016review}. For instance, Liu et al.  \cite{liu2012condition} developed a graphical software-based condition monitoring system for a Francis turbine. Different variables that had a cumulative impact on one another and affected the turbine’s performance were recorded in a database. These variables include pressure fluctuation, cavitation, vibrations, and turbine efficiency. The signals recorded from these sensors were then processed using Wavelet Analysis mainly due to its less computational requirement and complexity without affecting the system efficiency. The results segmented critical and dangerous operating zones and reflected the current global good state of the turbine. 

A vast majority of the predictive maintenance solutions for hydropower projects rely on data-driven methods. In \cite{wang2016review}, several data-driven predictive maintenance models for hydro turbines are analyzed and evaluated. The paper classifies the models into three categories, namely (i) physical models, that analytically derive the relationship between the system condition and system performance, (ii) stochastic models, that are based on probabilistic and statistical analysis, and (iii) data-mining models, that utilize advanced ML algorithms, such as ANN, Composite Neural Networks, SVMs and Decision Trees for fault identification.  One such data-driven solution is provided by Selak et al. \cite{selak2014condition}, where an SVMs-based condition monitoring and fault diagnostics system for hydropower plants. Data was acquired from a control system along with the high-frequency signals recorded from a dedicated data acquisition system. Around 108 features were extracted from the acquired data. The stored data was then transmitted to a Virtual Diagnostic Center (VDC) where the data was classified using SVMs and faults were diagnosed. SVMs exhibited a very high training accuracy as compared to other classification methods. The main shortcoming of SVMs is its higher computational time. However, this drawback can be addressed by reducing the training data and opting for parallel computation. 

Some methods also rely on some feature selection methods to reduce the features set by extracting the most relevant information. Some most commonly used methods to this aim include principal component analysis (PCA) and independent component analysis (ICA). Since condition monitoring in hydro turbines is typically non-linear, non-gaussian and a multivariable process, the conventional models based on PCA and ICA are not very effective. Hence, Zhu et al.  \cite{zhu2014novel} proposed a novel Kernel Independent Analysis (KICA-PCA) fault detection model with cumulative advantages of both PCA and ICA. Initially, the acquired non-linear process of monitoring data is transformed to higher dimensional linear data by using a kernel function. Features are then extracted from the vibration data by using ICA. The vibration faults are detected using Hotelling’s Statistics T2 and Squared Prediction Error (SPE). The results obtained from a real-life experiment reflected that KICA-PCA outperformed both PCA and ICA-PCA in terms of detection delays and detection success rate. 

Cloud computing has also been utilized to fulfill the computational requirements of predictive maintenance of hydropower projects. Kande, M. et al. \cite{kande2017rotating} provides a detailed overview of in-practice deployment methods that support the integration of monitoring systems either physically at the site or remotely through cloud or internet of things (IoT) devices in the industrial automation systems. These techniques can, however, be further analyzed from the perspective of security. As the future CBM systems for rotating electrical machines shall be cost-effective, having enhanced security and capability of large-scale deployment. Moreover, the cloud deployments of such systems may also result in time delays and network congestion. To avoid such issues, Xiao et al.  \cite{xiao2018prognostics} proposed a new prognostics and health management system for hydropower plants based on fog computing and docker container. The proposed fog computing technology overcame the problem of time delays and network congestion and improved the real-time processing capability of a cloud architecture. Features, called Health Indices, were calculated through a proposed storm-based data flow processing method. Furthermore, a distributed micro-service for seamless integration of different equipment and a docker container architecture for fault prediction and health management of hydropower plants was also proposed. Finally, the proposed fog-computing-based fault diagnostic system was implemented in the Zhen Touba Hydro Power Plant. 

There are also some deep learning-based solutions for predictive maintenance of hydropower projects. For instance, Zhang, C. et al. \cite{zhang2019deep} proposed a deep neural network Multi-Scale Convolutional Recurrent Encoder-Decoder (MSCRED) for unsupervised anomaly detection and diagnosis in multivariate time series data. The multiple levels of the various system statuses in distinct time steps were first characterized by constructing multi-scale signature matrices. Based on the signature matrices, inter-sensor correlations were encoded using a convolutional encoder. The temporal patterns were then captured by using an attention-based Convolutional Long-Short Term Memory (ConvLSTM) network. Finally, the temporal patterns and feature maps were input to a convolutional decoder to reconstruct the input signature matrices. Anomalies were further diagnosed by using the residual signature matrices. Finally, the experiments conducted on a synthetic power plant dataset demonstrated that the proposed MSCRED performed better than the state-of-the-art baseline prognostic methods.  

Some predictive maintenance methods also aim at fault detection in the bearing and gearbox of the hydropower project, where vibration data is analyzed. For instance, Wang, et al. \cite{wang2017prognostics} provided a detailed review of vibration-based bearing and gear health Indicators. The authors under the framework of prognostics and health management explained the relationship between the health indicators and the remaining useful life. Based on gear run to failure and accelerated bearing degradation data, the authors reviewed the vibrations-based health indicators and their associated problems along with some suggestions for future work to further enrich the already reviewed methods.
 Pino, G. et al.  \cite{pino2018bearing} devised a system to diagnose the deterioration in the guide bearings of Hydro Power Plants by using wavelet packet transform and  HMM with orbit curves. The start and stop frequency of a medium-sized HPP is increased while operating at peak loads. This increases the degradation of the mechanical components as the turbine is forced to operate in the transient state rather than the nominal condition. This operation in the transient state increases the extent of mechanical vibrations in the bearings installed in the runner. The vibration readings recorded during start-ups were filtered by a mean value DC component and wavelet packet transform. Based on the filtered data series, a relationship between the maximum displaced orbit curves and the operating hours was estimated. Transition probability distribution and observation probability distribution were calculated and both the matrices were provided as an input to the Hidden Markov Model (HMM) to simulate the degradation process and obtain the sequence of maximum orbit curve displacements. The system proved to be very effective in reducing the failure risk and maintenance costs as it calculated the deterioration level probabilities of the guide bearings. 

The above discussion reflects that adequate research has been conducted on the prognostics and condition monitoring of hydropower projects. However, the variation in the data of hydropower plants running under different operating conditions is quite extensive, thereby making the adaptability of the existing systems a very complex process. Furthermore, the real-life operational data specifically the equipment failure data for the hydropower plants are not easily available. Hence, most of the available systems are trained on synthesized data generated in a lab. Therefore, a robust and resilient prognostic maintenance system that can be implemented as a real-life application in a hydropower plant is still in infancy and requires further research.

\subsection{Wind Power Systems}
To increase the plant availability of the large wind turbines whether off-shore or on-shore, a cost-effective condition-based monitoring system is required because of the following reasons \cite{tavner2007reliability,tavner2008reliability,tchakoua2014wind}:

\begin{itemize}
    \item Since the capital cost associated with the wind turbines is quite high, and increased plant availability eventually improves the payback period.
\item The failure probability of large wind turbines is higher since they operate on a constantly variable load and are more prone to extreme environmental conditions.
\item Failure downtimes of large turbines are higher due to access difficulties.
\end{itemize}

Hence, an improved condition monitoring system prevents the subassemblies of the wind turbine from getting critically damaged, thereby reducing plant downtime.

To this aim, in the literature, several interesting solutions have been proposed. For instance, Yang et al.  \cite{yang2009intelligent} proposed a condition-based monitoring system for large wind turbines with the help of  Empirical Mode Decomposition (EMD) by keeping in view its advantages in dealing the non-stationary and non-linear signals. Experiments conducted on a specially designed drive train test rig reflected that the proposed system efficiently detected both the mechanical and electrical faults. However, their results are solely based on the analysis of generator power signal, whereas, the vibration and torque signals are only recorded for the purpose of verification. 

The transfer of power between the turbine and the generator shafts of an indirect drive wind turbine is being carried out by one of the most important subsystems known as a gearbox. The capital costs of wind turbines are quite high whereas, the gearbox alone contributes to a massive 13\% of the total cost of a 5MW wind turbine \cite{european2009economics}. Likewise, the maintenance and repair cost of gearboxes is also quite high. The gearbox of wind turbines is also prone to several failures. Smolders et al. \cite{smolders2010reliability} estimated the failure rates of critical components of a gearbox using the formulae and data available at NSWC-07 standard (Handbook of Reliability Prediction Procedures for Mechanical Equipment) issued by the Naval Surface Warfare Center Carderock Division on July 31, 2007. The failure rates of three generic gearbox configurations were predicted using a tool namely \textit{Relex Software} \cite{smolders2010reliability}. Similarly, a detailed review of the condition monitoring and fault diagnosis technologies used for the wind turbine gearboxes has been carried out by Nie et al. \cite{nie2013review}.

To reduce the O\&M costs and improve the wind turbine availability, a cost-effective condition monitoring technique has been also proposed by in \cite{yang2009cost}. Comparing to the conventional system analysis, the proposed technique has a reduced capital cost, detected both the electrical and mechanical faults, and applicable to both direct-drive and geared wind turbines. The newly proposed energy tracking methodology based on the adaptive Continuous Wavelet Transform (CWT) significantly reduced the calculation time for feature extractions from lengthy signals. Hence, realizing the possibility of a fast and lesser time-intensive condition monitoring.

Moreover, experiments and case studies have mostly been conducted on the condition monitoring of wind turbines for detecting failures and faults based on the alarms generated by Supervisory Control and Data Acquisition  (SCADA) module. However, very little work is carried out on the optimization of the SCADA alarm system. To this aim, Qiu et al. \cite{qiu2012wind} conducted an analysis on the wind turbine SCADA alarm system for improving its reliability. A current Oil \& Gas industrial standard was used as a benchmark to evaluate the performance of a wind turbine alarm system. Probabilistic and time sequence methods were proposed and tested on the wind turbine’s converter and pitch subassemblies. The case studies revealed complex patterns in the alarms arising from these subassemblies, that were then used to identify root causes of the failures.

Pertinent from the studies reported in the literature, it is evident that the wind turbines are operated under extreme environmental conditions, and the total operational unavailability over the entire life of a wind turbine reaches 3\%. Moreover, the contribution of operational costs of wind turbines to the total cost of energy (COE) for a wind power project is around 10\% to 20\%, initially, that can reach to an excess of 35\% towards the end of their life-timeline. We note that the designed operational life of wind turbines is around 20 years \cite{echavarria2008reliability,hahn2007reliability}. An effective preventive maintenance strategy can hence lower the frequency of shutdowns thereby reducing the operational costs \cite{hines2013sandia,walford2006wind,fischer2011reliability} and increasing the RUL of the wind turbine. A state-of-the-art reliability-centered maintenance strategy including the preventive maintenance and condition monitoring of the subsystems is therefore required, which may help to determine an optimum maintenance point between the corrective and scheduled maintenance strategies \cite{fischer2011reliability,mcmillan2008condition,amayri2011condition,besnard2010approach,tchakoua2014wind,yang2014wind}.

One of the most important components of a wind turbine condition monitoring and fault diagnosis system is the acquired signals along with the processing methods applied over these signals. The signal processing is based on various distinct methodologies including the classical time-domain analysis methods  (statistical analysis, Hilbert transform, and envelope analysis),  the classical frequency analysis methods like Fast Fourier Transforms (FFT), the time-frequency analysis methods like wavelet transforms, and the probability and model-based approaches like the Bayesian and AI methods. Qiao et al.  \cite{qiao2015survey} reviewed and compared the capabilities, functionalities, and limitations of these methodologies.

The diagnosis of mechanical faults in wind turbines cannot be efficiently carried out by merely analyzing the power signals. Instead, evaluation of vibration signals acquired from the accelerometers helps to improve the process of successful diagnosis. Santos and Villa et al. \cite{santos2015svm} presented a fault diagnosis system for the wind turbines based on the combination of multi-sensory data with a data mining solution. The vibration signals acquired from the accelerometers were processed using the angular resampling technique and the classification was carried out using SVMs. The system was validated on a testbed simulation of a wind turbine having misalignment and imbalance as fault topologies. The performance of SVMs and ANNs were compared and it was found that the linear kernel SVMs outperformed ANNs in terms of accuracy, training, and tuning times.

A key feature in planning the industrial operation and maintenance activities is the detail of the progression of the faults along with the RUL of the equipment. Herp et al. \cite{herp2020assessment} presented models for the estimation of RUL of wind turbine main bearing using the likelihood functions based on the very concepts of health management, prognostics, and survival analysis. This study, after a thorough analysis of 67 different wind turbines, concluded that they discontinue functioning 13 days before their failures thereby accumulating a total of 786 days of potential non-operations. Similarly, Bill et al. \cite{lau2012pecht} summarized the most effective and efficient prognostic methodology adopted by various researchers against the specified turbine fault, depicted in Table \ref{tab:methods} below.

\begin{table}[]
\centering
\caption{Matching of Fault Prognostic Method and Turbine Failures \cite{herp2020assessment}.}
\label{tab:methods}
\begin{tabular}{|c|c|}
\hline
\textbf{Failure} & \textbf{Prognostic Method(s)} \\ \hline
Electric Control &  Hidden Markov Model\\ \hline
 Grid& HMM \\ \hline
 Blades & HMM \\ \hline
  Coupling &  HMM\\ \hline 
  Hub &HMM  \\ \hline
  Tower  & HMM \\ \hline
   Foundation  & HMM \\ \hline
   Hydraulic   & Particle Filter \\ \hline
  Entire nacelle &  Particle Filter\\ \hline
  Entire turbine & Particle Filter \\ \hline
 Brake  &  HMM and NN \\ \hline
  Air brake & HMM and NN \\ \hline
 Mech. Control  &HMM and Particle Filter \\ \hline
 Generator  & NN and Particle Filter \\ \hline
 Gearbox  &  All Three (03) methods\\ \hline
 Yaw system   & All Three (03) methods \\ \hline
   Axle/Bearing   & All Three (03) methods \\ \hline
\end{tabular}
\end{table}

As evident from the above discussion, extensive research has been conducted in the area of CBM of wind-powered turbines, but quantification of faults and prognostics are still at a nascent phase. The accuracy assurance of the prognostics methods and the metrics of qualification as described by Saxena et al. \cite{saxena2008metrics} are hence required to be carried out. Multiple attempts have been made to determine the equipment damage using both the data-driven and model-based approaches. However, keeping in view the complexity of the problem and the time extensive task of simulating the failures, no single standard approach has been reliably proven to perform consistently well. Furthermore, analyzing the effects of external factors on the RUL predictions of wind turbines is still an open question. Hence, before the incorporation of the results into the decision-making process for conducting the maintenance, the reliability of the developed prognostic methods and systems should be thoroughly analyzed and assessed. 

\subsection{Solar Power Systems}
Solar-powered generating sources being purely electrical systems and are less prone to degradation and faults as compared to the electro-mechanical systems, which comprise rotating parts including the bearings. Hence, CBM for solar power projects is still an unexplored area of research. Therefore, the relevant literature in this field is limited. However, some of the studies that are being carried out pertinent to CBM of Solar Powered systems have been discussed here.

The monitoring of Photovoltaic (PV) systems normally incorporates and measures the module and array performance, stability of the grid, power factors, and prevention of islanding. However, studies are now being conducted for the development of PV prognostics and health management systems to mitigate the arc-fault and ground-fault. The PV monitoring systems are hence expected to detect the faults when the system behavior deviates from the normal operational condition \cite{chouder2010automatic,chao2008modeling, karatepe2011controlling,stettler2006spyce,drews2007monitoring,houssein2010monitoring,yagi2003diagnostic}. To this aim, various models have been designed to calculate the expected power based on the temperature and irradiance data gathered either from the sensors \cite{chouder2010automatic,chao2008modeling,karatepe2011controlling} or the weather and satellites \cite{stettler2006spyce,drews2007monitoring}, which is then compared against the actual valves to analyze the changes in the behavior of the system, if any. 

Riley et al. \cite{riley2012photovoltaic} designed an ANN-based PV prognostics and health management system for monitoring the system health, measuring degradation, and generate maintenance schedules. The system when tested on research installations effectively detected system faults that caused a reduction in the output power.
The Potential Induced Degradation (PID) affected silicon solar modules exhibit a significant power loss even at new PV installations. Hence, thermographic imaging can be used as a quick and reliable tool for detecting PID-affected modules. Kaden et al. \cite{kaden2015power} presented a thermographic image-based prognostic system to detect the power loss in the PID-affected silicon solar modules without dismounting them by using remote controlled multicopters. Karoui et al. \cite{karoui2018diagnosis}, on the other hand, carried out the performance analysis of energy storage systems under the umbrella of the prediction of possible degradation. To support the proposed approach, they developed and tested an Advanced Data Analysis Tool and named it “A4”. This tool proved to be effective in the operation aiding of the installed systems and helped in improving the performances and availability. 

The economic viability of micro-grids having distinct and distributed energy sources can only be ensured by improving their reliability and reducing maintenance costs. Hence, the need for a real-time condition monitoring and prognosis health system is inevitable. To this aim, Feng et al. \cite{feng2006survey} proposed an experimental and non-invasive online Condition Monitoring and Health Prognosis System (CMPHS) for an isolated microgrid testbed. Real-time condition monitoring of PV modules and cell-level monitoring of the storage batteries was carried out. Data was transmitted and stored in a database for visualization and carrying out prognosis by applying data analytics and data mining techniques. The prognostics helped increasing system performance, reducing maintenance costs, and the risk of grid instability.

The ratio of the actual energy generated to the rated generation capacity of the solar cell is known as the Performance Ratio (PR) \cite{dierauf2013weather}. PR of the solar cell is one of the key indicators of the reliability of a solar PV system. The prediction of the solar cell performance enables the energy providers in taking critical decisions on carrying out plant maintenance and replacement of solar cells that are having degraded performance efficiency. Bandong et al. \cite{bandong2019performance} improved the energy reliability by estimating and predicting the PR of Solar Power Plants using machine learning techniques. A principal component analysis and SVMs (PCA-SVMs) based model was designed to estimate the PR by using three years of weather data. The results were then compared to a Multiple Linear Regression (MLR) based prediction model. Evaluation of the results based on the Root Mean Square Error (RMSE), the coefficient of determination (R2) and Mean Absolute Percentage Error (MAPE) showed that the PCA-SVMs outperformed MLR in terms of accurately predicting the PR.

Deep learning has also been used for the prognosis of solar energy systems. For instance, Correa et al. \cite{correa2020assessment}  assessed several deep learning techniques including ANN, recurrent neural networks (RNN), and Long Short-Term Memory (LSTM) for the prognosis of solar thermal systems. The analysis of the results revealed that LSTM outperformed DNN in terms of accuracy and yielded more precise values than RNN for temperature sequence predictions. However, all three models predicted the temperature sequences better than the naïve persistence forecasts and other regression techniques, such as Bayesian Ridge, Gaussian process, and Linear Regression.

In contrary to the previous that was mostly based on the simulated data, Kelker et al. \cite{kelker2019development} relied on real measurement data of a local PV system for successfully developing a forecast model to predict the output power of a photovoltaic system using feed-forward artificial neural networks. 

Numerous models for the PV systems have been developed and employed including the specific fits models \cite{drews2007monitoring}, the PV circuit models \cite{chouder2010automatic,houssein2010monitoring}, matter-element models \cite{chao2008modeling} and the state of the art systems with an updated warning criteria \cite{yagi2003diagnostic}. The models that are based on the data acquisition from the physical systems including current, voltage and power measurements have been incorporated to detect various fault conditions such as shading \cite{chouder2010automatic,stettler2006spyce,drews2007monitoring,houssein2010monitoring,yagi2003diagnostic}, the inverter failures \cite{stettler2006spyce,drews2007monitoring,yagi2003diagnostic}, snow covers \cite{stettler2006spyce,drews2007monitoring} short circuiting and failure of the modules \cite{karatepe2011controlling,houssein2010monitoring,yagi2003diagnostic} and malfunctions at the string-level \cite{stettler2006spyce,drews2007monitoring,chouder2010automatic}. Fault diagnoses and estimation of PV output has also been successfully performed by using the machine learning algorithms \cite{karatepe2011controlling,yagi2003diagnostic,al2000application}, Bayesian Networks \cite{coleman2011intelligent} and fuzzy logic \cite{elshatter2000fuzzy,abdulhadi2004neuro}. A method was devised for the tracing of Current-Voltage (I-V) curves to detect the degradation of the PV modules \cite{hamdaoui2009monitoring} but it was also not practically implemented at field installations. 

\section{Taxonomy of ML Techniques for Prognostic Maintenance}
\label{sec:taxonomy_ML}
In this section, we discuss some commonly used ML techniques for predictive maintenance of renewable energy. For simplification, we categorize the existing methods based on the nature/family of the ML algorithms.  

\subsection{Classical ML Techniques}
A vast majority of the works on predictive maintenance rely on classical ML techniques, such as SVMs, Random Forest (RF), Decision Trees, and Nearest Neighbours \cite{borunda2016bayesian}. There are several key characteristics of classical ML algorithms that make them a feasible solution for several tasks. For instance, these algorithms can be deployed on low-end machines with limited computational and storage capacities. However, the most key characteristic of these algorithms is high interpretability, which is a desirable feature in several sensitive applications, such as healthcare, predictive policing, education, and transportation \cite{adadi2018peeking}. However, these algorithms are suitable for small datasets only. Moreover, these algorithms require engineered features, which are very hard to extract and generally require a deep knowledge of the domain.

Most of the initial efforts for predictive maintenance of renewable energy systems rely on classical ML algorithms trained on engineered features \cite{mohandes2004support}.  In most of the works, SVMs are used for classification purposes. For instance, Kusiak et al. \cite{kusiak2011prediction} employed several classical ML algorithms including Boosting Tree Algorithm (BTA), SVMs on features extracted from wind turbine data. The feature set is composed of a total of 60 different parameters/features categorized into four groups including wind, energy conversion, vibration, and temperature parameters. Sugumaran et al. \cite{sugumaran2007feature} also used SVMs along with decision trees as a classifier for fault diagnostics of roller bearing. In  \cite{widodo2007combination}, SVMs are trained on features selected through independent component analysis (ICA) for fault monitoring, detection, and classification of the faults that occurred in induction motors. Besides SVMs, some other classical ML algorithms, such as Bayes classifier, Decision Trees, and RF are trained on engineered/hand-crafted features \cite{orchard2007particle,sugumaran2007feature,dhibi2020reduced}.

\subsection{Fuzzy Logic}
Fuzzy Logic is the type of AI algorithms that aims to achieve intelligence by creating fuzzy logic of parameters involved in the application \cite{alakhras2020survey,feng2006survey}. One of the key advantages of fuzzy logic-based methods is the high interpretability as the outcome of these methods is understandable by a human. The ability to be highly interpretable is based on its working mechanism as the fuzzy classes and rules are defined by human experts. However, defining such rules require a deep knowledge of the domain, which is also one of its key limitations compared to other types of AI techniques, such as ANNs.  

Similar to other application domains, such as communication \cite{arif2020optimization}, healthcare \cite{jayalakshmi2021fuzzy}, and industrial production \cite{caiado2021fuzzy}, fuzzy logic-based methods have been widely employed for predictive maintenance of renewable energy systems \cite{suganthi2015applications}. For instance, Merabet et al. \cite{merabet2015condition} propose a fuzzy logic-based framework for condition monitoring and fault detection in wind turbines. In \cite{qu2020wind}, a novel mechanism of expanding linguistic terms and rules in the fuzzy logic system is proposed for fault detection in wind turbines using SCADA data. In \cite{adhikari2016fuzzy}, on other hand, a fuzzy logic-based solution is proposed for fault detection and classification in power transmission lines. Other interesting works relying on fuzzy logic-based solutions for predictive maintenance of renewable energy systems include the works presented in \cite{abdali2017fast,noureddine2018fuzzy,zaki2019fault,hichem2017fuzzy,kumar2018power}.

\subsection{Hidden Markov Models (HMMs)}
HMMs belong to the generative models family of ML that allows to model and describe the evolution of observable events depending on hidden or unobservable internal factors.  In the literature, HMMs have been widely explored for a diversified set of applications, such as speech recognition, speech-to-text translation, modeling and analyzing DNA sequencing errors, and social interactions \cite{adams2019survey}. There are several variants of HMMs, such as profile-HMMs, pair-HMMs , and context-sensitive HMMs \cite{cappe2006inference}. 

There are several advantages of HMMs that make it a better choice for predictive maintenance of renewable energy systems. For instance, HMMs are very effective in fault diagnosis and system degradation analysis of random and dynamic systems and non-stationary signals particularly in the case of systems having multi failure modes. These algorithms also allow us to learn from incomplete data and are able to model both temporal and spatial data.  

There are also some limitations of HMMs. For instance, these models do not work for scenarios where the failure states are observable. Moreover, extensive training data is required for accurate modeling. HMMs are better suited in scenarios where the number of hidden states is not very large and are computationally expensive in situations where the number of hidden states is large. Moreover, HMMs can only provide an estimation since the true state transitions cannot be observed. 

The literature covers some interesting predictive maintenance solutions relying on HMMs. For instance, \cite{kouadri2020hidden} proposed an HMM-based framework for fault detection in wind energy systems. Similarly, in \cite{shin2018development}, HMMs are employed to detect faults in mechanical parts of a wind turbine using vibration data obtained from a  3 MW wind turbine. In \cite{omoregbee2018fault}, a modified version of HMMs namely ``Bayesian robust new hidden Markov model (BRNHMM)'' is employed for fault detection in roller bearing where HMMs parameters are estimated through Bayesian inference. Besides fault detection in the energy systems, HMMs are also utilized for the detection and classification of faults in power transmission lines \cite{freire2019transmission}.

\subsection{Neural Networks (NNs)}
NNs are inspired by the human brain, and process data the way the human brain does. ANN refers to the network of artificial neurons designed for a particular application. Due to the outstanding self-learning capabilities and ability to work with insufficient knowledge, NNs have been widely utilized in a diversified set of applications of speech, images, text, and video processing/recognition \cite{ahmad2019deep,wu2020comprehensive,ahmad2021sentiment}. Some key advantages of NNs include the ability to model non-linear, multi-dimensional, complex systems. Moreover, NNs are robust towards noise due to the ability of processing non-linear information.

There are several types of NNs architectures, such as Feed-forward Neural Networks, Single and Multilayer Perceptron Neural networks, Recurrent Neural networks (RNNs), LSTM, Modular Neural Network (MNNs), and CNNs.

In the literature, several types of NNs have been employed for predictive maintenance of renewable energy systems \cite{ferrero2019review}. ANNs (i.e, feed-forward neural networks), RNNs, and CNNs are among the most commonly used architecture of NNs used for the predictive maintenance of energy systems. For instance, in \cite{rao2019solar}, feed-forward neural networks (FFNN) are employed for fault detection in solar energy systems. In \cite{kavaz2018fault}, an FFNN-based framework is proposed for the detection and classification of wind turbine sensor faults. There are some other interesting works relying on FFNNs for fault detection renewable energy systems \cite{erouglu2019early,martinez2018labelling,jiang2017wind}.   

In the RNNs family both LSTM and biLSTM have been widely utilized \cite{yin2020fault,chen2021anomaly,lei2019fault,han2021combination}. For instance, Xiang et al. \cite{xiang2021fault} employed LSTM with an attention-based mechanism for fault detection in wind turbines. Similarly, in \cite{choi2020consistency}, an LSTM-based framework is proposed for fault detection in nuclear power systems.  In \cite{zou2021bearing}, a bi-LSTM-based framework has been employed for bearing fault diagnostics. A bi-LSTM based solution is also proposed for fault detection in marine hydrokinetic turbines \cite{wilson2018bidirectional}.

CNN is another well-known architecture of NNs. Similar to other applications, such as video and image recognition and classification \cite{ahmad2019deep}, CNNs have also been widely explored in predictive maintenance of energy systems. CNN is more effective in image-based solutions for the detection and classification of faults in energy systems. For instance, Shin et al. \cite{shin2021ai} proposed a CNNs based framework for fault detection in the gearbox and bearings of wind turbines. In \cite{ulmer2020early}, a CNN model is trained over SCADA data for early fault diagnosis in wind turbines. There are also some other interesting solutions relying on CNNs for predictive maintenance of renewable energy systems \cite{jiang2018multiscale,mohammadi2020deepwind,helbing2018deep,ulmer2020cross,zhang2020mask,zare2021simultaneous,li2019fault}.

Table \ref{tab:properties_algorithms} summarizes the pros and cons of various prognostic maintenance techniques of renewable energy systems.

\begin{table*}[]
\centering
\caption{Comparison of Prognostic Maintenance Techniques for Complex Systems.}
\label{tab:properties_algorithms}
\scalebox{1}{
\begin{tabular}{{|p{2 cm}|p{7 cm}|p{7 cm}|}}
\hline
\textbf{Methodology} & \textbf{Advantages} & \textbf{Disadvantages} \\ \hline
Hidden Markov Model (HMM) & \begin{itemize}
    \item Performs effectively for fault diagnosis and system degradation analysis of random dynamic systems and non-stationary signals.
    \item Best suits the systems that are having multi failure modes.
    \item Ability to distinguish between various faults of different bearing types, if trained properly.
    \item Since the algorithm has the ability to model various stages of degradation, hence it can take into account non-monotonous failure trends.
    \item Ability to model both temporal and spatial data.
    \item Ability to quickly handle the incomplete data sets.
    \item The RUL predictions by HMMs also provides with the confidence intervals.
    \item Successful real-life implementations as the interpretation and understanding of the model is easy.
\end{itemize} &  \begin{itemize}
    \item The model does not work for scenarios where the failure states are observable.
    \item Proportionate to the hidden states, extensive training data is required for accurate modelling.
    \item The algorithm is computationally expensive for situations where the number of hidden states are large.
    \item The efficiency of the prognostics depends upon the selection of the failure threshold.
    \item Cannot model unanticipated and novel faults.
    \item The HMM can only provide an estimation since the true state transitions cannot be observed.
\end{itemize}\\ \hline
SVMs & \begin{itemize}
    \item High accuracy and robustness even with non-linear and higher dimension input data.
    \item Dynamic SVMs has got the better ability to model non-stationary time series.
    \item Extended decision boundaries.
    \item Higher accuracy and efficiency.
    \item Performs exceptionally well for both large and small datasets.
    \item Ability to carry out real-time analysis.
    \item Ability to be generalized without compromising the performance even against limited number of learning patterns.
    \item Works extremely well for carrying out fault diagnosis in the electro-mechanical systems and machines.
    \item Proves to be an effective prognostic method even for limited samples.
    \item SVMs is more memory efficient.
\end{itemize} & \begin{itemize}
    \item There are no set parameters for the selection of the kernel function.
    \item The training requires to be carried out for both positive and negative examples.
    \item The performance degrades for cases where the number of features exceeds the number of training data sample.
    \item There is no accurate probabilistic explanation to an SVMs model.
    \item It does not directly provide the RUL.
    \item Requires fine tuning of the parameters.
    \item The regression and classification outcome of SVMs are point estimates only.
    \item The problem of numerical stability may pop ups for constrained quadratic programming.
\end{itemize} \\ \hline
Artificial Neural Network (ANN) & \begin{itemize}
    \item Ability to model non-linear, multi-dimensional, complex systems.
    \item Robustness towards noise because of the ability to process non-linear information.
    \item Ability to carry out unsupervised learning and model complex phenomena. 
    \item Ability to efficiently tackle imprecise and complex data.
    \item The parallel model structure makes it computationally inexpensive and hence feasible for real-time operation.
    \item Ability to handle multivariate systems with speed and accuracy. 
    \item Ability to sustain network damage due to the resilient and fault tolerant behaviour.
    \item Provision of confidence intervals.
\end{itemize} & \begin{itemize}
    \item Extensive pre-processing of input data is required for dimensionality reduction and complexity.
    \item A large quantity of training data is required encompassing the variability and ranges and must be a true reflection of the original data.
    \item Model perfection can be a time consuming process.
    \item The process of decision making in trained network is left covert to the users.
    \item Failure diagnosis is based on the presumed threshold.
\end{itemize} \\ \hline
Fuzzy Logic & \begin{itemize}
    \item Ability to carry out condition based classification based on historic data.
    \item Ability to work effectively for complex systems and imprecise data.
    \item Performs exclusively in developing data models based on uncertainty. 
    \item Interpretation of the model is simpler.
    \item Provision of the confidence interval.
\end{itemize} & \begin{itemize}
    \item Unable to provide exact time and probability of failure.
    \item Performance degrades in cases where the member functions are complicated.
    \item Inability to self-learn and requires the help of  domain experts for the development of fuzzy rules.
\end{itemize} \\ \hline

\end{tabular}}
\end{table*}

\begin{table*}[]
\centering
\caption{Summary of some relevant works on Prognostic maintenance of renewable energy in terms of application/domain, type of ML technique used, and a brief description of the method.}
\label{videos_summary}
\scalebox{0.78}{
\begin{tabular}{|p{.2cm}|p{4.5cm}|p{4.5cm}|p{11cm}|}
\hline
\multicolumn{1}{|c|}{Refs.} & \multicolumn{1}{c|}{Application} 
&\multicolumn{1}{c|}{ML Technique} &\multicolumn{1}{c|}{Description of the Method} \\ \hline

\cite{sugumaran2007feature} & Faults diagnosis of induction motors &  Classical ML (SVMs, Decision Trees)  & Initially, ICA is employed for feature selections. AN SVMs and Decision trees classifiers are then trained on the selected features. Moreover, the fault detection of induction motors is treated as a multi-class classification task where the models have to differentiate among multiple fault classes, such as bearing fault, bowed rotor, mechanical unbalance, misalignment, broken rotor bar, etc. \\ \hline

 \cite{santos2015svm}& Fault Detection in Wind Turbines
 &  Classical ML (SVMs)& Proposes a multi-modal solution where data from multiple sensors from wind turbines is analyzed through data mining techniques for potential part detection wind turbines. SVMs with three different types of kernel functions are employed for classification purposes. \\ \hline

\cite{qu2020wind} & Fault detection in wind turbines & Fuzzy Logic & Relies on expanded linguistic terms and rules using non-singleton fuzzy logic. Moreover, four different types of experiments were conducted on real-time wind turbine data to evaluate the proposed solution.  \\ \hline
 
 \cite{zaki2019fault} & Fault detection Solar power systems &  Fuzzy Logic& Proposes fuzzy logic control (FLC) method for detection and classification of eight different types of faults in solar systems. To this aim, the measured values/readings of several electrical parameters are compared against their theoretical valves.  \\ \hline
  
\cite{omoregbee2018fault}   & Fault detection in bearing & HMMs & Proposes a modified version of HMMs namely ``Bayesian robust new hidden Markov model (BRNHMM)'' for fault detection in roller bearing where HMMs parameters are estimated through Bayesian inference.  \\ \hline
  \cite{kouadri2020hidden}  & Fault detection in wind turbine & HMMs &  Initially, Principal Component Analysis (PCA) algorithm is used to extract key features from the wind turbine data. A HMMs-based framework is then used on the extracted features for fault detection. The features extracted through PCA are treated as observable in the HMMs-based solution.\\ \hline
  \cite{rao2019solar}  & Fault detection in solar power systems & ANNs (FFNNs) & Proposes a FFNN-based framework to diagnose faults in photovoltaics array using data obtained through different sensors including current, voltage, and temperature readings. \\ \hline
  \cite{kavaz2018fault}  &  Fault detection in wind turbine sensors& ANNs (FFNN) & Proposes a FFNN based framework for fault detection in wind turbine sensors using temperature data. The basic motivation behind the method is to avoid errors in data due to faulty behaviour in sensors. \\ \hline
 \cite{xiang2021fault} &Fault detection in wind turbine  & ANNs (LSTM) & Proposes a hybrid framework for fault detection in wind turbine, where CNNs are cascaded to LSTM based on attention mechanism. CNNs are trained on SCADA data while the attention mechanism is applied to give more importance/weight to important information in LSTM.  \\ \hline
 \cite{zou2021bearing} & Fault detection in bearing & ANNs (bi-LSTM) & Proposes a framework to combine multi-scale weighted entropy morphological filtering (MWEMF) signal processing and bi-LSTM for fault detection in beaning.  \\ \hline
 \cite{shin2021ai} & Fault detection in gearbox and bearing of wind turbine & ANNs (CNNs) & Proposes a CNNs based framework for fault detection in endoscope images. The CNNs architecture/model is composed of four convolutional and two fully connected layers. \\ \hline
 \cite{ulmer2020early} & Fault diagnosis in wind turbine & ANNs (CNNs) &  A CNNs model composed of  four convolutional and two fully connected layers is trained on SCADA data for early diagnosis of faults in wind turbine. \\ \hline

\end{tabular}}
\end{table*}

\section{Benchmark Datasets}
\label{sec:datasets}
Predictive maintenance not only results in a significant reduction in maintenance cost but also increases the lifetime and productivity of different components of power plants. However, the feasibility and effectiveness of predictive maintenance are largely dependant on the availability of quality data to train and evaluate predictive models for monitoring, detection, and forecasting faults. Being one of the most crucial aspects of predictive maintenance, in this section, we provide a detailed overview of some of the publicly available datasets in the domain.

\begin{itemize}
    \item \textbf{Endoscope Bearing Images Dataset \cite{shin2021ai}}: The dataset is composed of 2,301 endoscope images and aims to facilitate predictive maintenance by fault diagnostics of wind turbines. All the images are captured during inspections of wind turbines, where wind turbine gearboxes and main bearings of 138 plants are analyzed. One of the key characteristics of the dataset is the diversity in terms of models' manufacturers as well as the operating age of the components. Moreover, the images are also taken with various types of endoscopes resulting in variation in image resolution, illumination, and quality. The training set is composed of 2,101 while the test set contains a total of 210 endoscopic images. The images are categorized into two classes namely (i) \textit{normal}, and \textit{abnormal} images, where each image is analyzed by at least three experts, and the final decision is made based on majority voting.  
    
    \item \textbf{Vibration Signal from Wind Turbine Dataset \cite{martin2018dataset}}: This dataset also aims at monitoring wind turbines for potential fault detection via vibration signals from wind turbines. The dataset contains a total of 16,384 signals generated by six different but nearly located turbines each possessing a three-stage gearbox. The data is collected from the axial direction of an accelerometer at a sampling rate of 12.8 kilosamples/second. Moreover, the vibration signals are segmented uniformly resulting in a collection of segments each 1.28 seconds long. The dataset is organized into six files each containing vibration signal segments of a separate wind turbine. The features/information covered in the dataset includes (i) the number of years for which the vibration data is recorded, (ii) the speed of the turbine in terms of cycles per second, and (iii) the vibration signal time series expressed in Gs. 
    
    \item \textbf{Bearing Data Set \cite{NASA_DATASET}}: The data set is collected at the Center for Intelligent Maintenance Systems (IMS), University of Cincinnati. To this aim, several test-to-failure experiments were conducted where four different bearings were installed on a shaft under a load of 6000 lbs. The dataset is further divided into three different subsets, each containing data related to a separate test-to-failure experiment. These subsets are composed of vibration signal segments each of one-second duration. Moreover, each segment is sampled at 20 kHz and contains 20,480 different data points. No separate training and testing sets are provided.  
    
    \item \textbf{Wind Turbine Gearbox Dataset \cite{Bearing_datasets,sheng2012wind}}: The dataset is collected by National Renewable Energy Laboratory (NREL) using a customized vibration data acquisition system. The data acquisition system is composed of a total of twelve accelerometers to record low-speed shaft torque and generator speed along with the accelerometer data. The dataset provides vibration data related to two different classes, namely (i) healthy, and (ii) damaged gearbox. All the data samples are taken from the same design of wind turbine gearbox. During the data collection, the damaged gearboxes are tested in both dynamometer and wind farm where the test results are recorded for ten minutes under each test condition. The data collected under each test condition is then organized into 10 files, resulting in a total of 30 files.  

\item \textbf{Wind Generator Dataset \cite{mammadov2019predictive}}: This dataset also aims at facilitating AI-based predictive maintenance of wind turbines by provides data from four different turbines. Each sample is composed of 101 different features as detailed in \cite{mammadov2019predictive}. Some key features of wind turbines covered in the dataset include environmental conditions, such as operational time and average and standard deviation of wind speed, measurements for wind turbine components, such as average rotations per minute, and electrical variables, such as voltages, currents, and frequency. All the data samples are labelled either as \textit{faulty} or \textit{normal}.   
\end{itemize}

Table \ref{datasets_videos} provides a summary of some of the publicly available datasets for training and evaluation of predictive maintenance algorithms for renewable energy systems. 

\begin{table*}[]
\centering
\caption{Summary of the datasets prognostic Maintenance.}
\label{datasets_videos}
\scalebox{0.78}{
\begin{tabular}{|p{.2cm}|p{3cm}|p{4.5cm}|p{8.5cm}|}
\hline
\multicolumn{1}{|c|}{\textbf{Refs.}} & \multicolumn{1}{|c|}{\textbf{Application}} & \multicolumn{1}{c|}{\textbf{Statistics}} 
& \multicolumn{1}{c|}{\textbf{Notes}} \\ \hline \hline
\cite{martin2018dataset} & Fault detection in wind turbines & 16,384 samples; sampling rate is 12.8 kilosamples/second; signal segment length is 1.28 seconds & The data is collected from six different nearly located wind turbines. The features/information covered in the dataset include: (i) number of years for which the vibration data is recorded, (ii) speed of the turbine in terms of cycles per second, and (iii) the vibration signal time series expressed in Gs. \\ \hline
 \cite{NASA_DATASET} & Fault detection in Bearing & 20,480 samples/data points; sampling rate is 20 kHz &  The dataset is divided into three different subsets, each containing data related to a separate test-to-failure experiment. These subsets are composed of vibration signal segments each of one-second duration. No separate training and testing sets are provided\\ \hline
 \cite{Bearing_datasets,sheng2012wind} & Fault detection in wind turbine gearbox & 30 files in total; 3 different test conditions; 10 files per test condition  & The dataset provides vibration data related to two different classes, namely (i) healthy, and (ii) damaged gearbox. All the data samples are taken from the same design of wind turbine gearbox. The data collected under each test condition is then organized into 10 files, resulting in a total of 30 files.  \\ \hline
   \cite{mammadov2019predictive}  & Fault detection in Wind turbine generator & 2 classes: \textit{faulty} or \textit{normal}; 101 different features & Some key features of wind turbines covered in the dataset include environmental conditions, such as operational time and average and standard deviation of wind speed, parameters of wind turbine components, such as average rotations per minute of a generator, and electrical variables, such as voltages, currents, and frequency. \\ \hline
   \cite{shin2021ai}    & Fault detection in bearings  &  2,301 samples; training set contains 2,101 while test set includes 210 endoscopic images; 2 classes \textit{normal} and \textit{abnormal} & All the images are captured during inspections of wind turbines, where wind turbine gearboxes and main bearings of 138 plants are analyzed. The images are also taken with various types of endoscopes resulting in variation in image resolution, illumination, and quality.  \\ \hline
\end{tabular}}
\end{table*}

\section{Challenges, Open Issues, and Future Research Directions}
\label{sec:challenges}

\subsection{Challenges and Open Issues}

The most common issues and challenges in predictive maintenance of renewable energy systems are:  

\begin{itemize}
    \item \textbf{Availability of Data}: The feasibility and performance of ML models especially deep learning in any application is largely constrained by the availability of quality data to train and evaluate the models \cite{ahmad2020developing}. Predictive maintenance of renewable energy plants is one of the domains lacking in quality training data. Since the requirement of data for the data-driven approaches is extensive, due to the unavailability of quality data especially for the newly designed systems, in most cases it is not possible to deploy data-driven approaches in the domain. Moreover, over-fitting and over-generalization during the training of the algorithm due to the unavailability of quality data can also affect the results. 
    \item \textbf{Data Auditing}: Data auditing, which is the process of analyzing the quality and feasibility of the data for a particular application, is very critical in predictive maintenance of energy plants where noise generated by the mechanical parts is accompanied by the measurements. Therefore, it is important to analyze the measurements/data to identify and sustain multivariate and noisy data. Moreover, it is also recommended to identify the risks associated with the noisy data and the potential impact on the performance of the predictive models. To this aim, frequent tests/audit needed to be carried out to ensure the validity of the algorithms over a period of time, and they learn what they are intended to learn \cite{AI_auditing}.
    \item \textbf{Feature Engineering/selection}: Feature engineering/selection is one of the key phases of predictive maintenance of renewable energy systems where key features (i.e., variables) relating to the outcome of the models are identified. This process is very challenging and is generally conducted manually or automatically under the supervision of an expert.  The extent to which the adopted approach needs to have an understanding of the technical domain knowledge, underlying failures, and data collection methodology, varies from model to model. The application of Neural Networks does not require a very detailed understanding of the failure processes whereas, the Expert and Fuzzy systems require a medium amount of understanding. On the other hand, the physical models require extensive domain knowledge and a detailed understanding of the environmental factors and physical mechanisms that influence the failure processes. 
    \item \textbf{Simultaneous modeling of Multiple Faults/Failures}: Simultaneous modeling of multiple failures can be computationally very intensive since it involves the processing of large datasets that affect the real-time performance of the algorithm. This challenge can however be addressed by acquiring suitable resources and a well-designed algorithm.
    \item \textbf{Design of Robust Models}: At times the results cannot be based on intuition because of the absence of the physical knowledge of the system or the novelty of the faults that have occurred. Therefore, it is important to develop models that are robust enough to handle and manage novel faults.
    
    \item \textbf{Trade-off between Performance and Explainability}: Explainability, which is one of the desirable characteristics of ML algorithms, refers to the ability of ML models to justify their predictions/decisions. Explainability is not always a desirable characteristic or key concern for the researchers rather the emphasis is on the performance of the models. However, we believe, predictive maintenance of renewable energy systems is one of the critical applications where prediction/decisions of ML models should be explainable and make sense to a human. Though a vast majority of the literature relies on traditional ML techniques and rule-based methods for predictive maintenance of the energy systems, some recent works are relying on deep learning-based methods. The traditional ML algorithms and rule-based methods are explainable but less accurate in making predictions. On the other hand, deep learning has been proved very effective in terms of prediction, however, they are black-box methods and do not provide an explanation of the prediction. It is very challenging to keep a balance between performance and explanation of the decision made in predictive maintenance of energy systems. 
   
   \item \textbf{Adversarial Attacks}: ML models for different applications in smart grids and energy systems including predictive maintenance models are also subject to adversarial and security attacks \cite{niazazari2020attack,mode2020adversarial}.  For instance, a vast majority of the methods rely on deep learning, cloud computing, and the Internet of Things (IoT) devices for predictive maintenance of energy systems. Both deep learning techniques and IoTs deployments are prone to adversarial and cyber-security attacks \cite{mode2020crafting}. 
    
\end{itemize}

\subsection{Future Research Directions}
Some key potential research directions in predictive maintenance of renewable energy systems are:

\begin{itemize}
    \item \textbf{Trad-off between Performance and Explainability}: As detailed earlier, most of the initial efforts for predictive maintenance of renewable energy systems rely on classical ML algorithms trained on hand-crafted features. The outcome of these algorithms can be easily understood. However, recently the trend shifted towards ANNs-based methods, which improved the accuracy/performance of the predictive maintenance techniques significantly. However, it is hard to understand the causes behind the decisions/predictions made by these methods. Considering the risks involved with the energy systems explainability seems a desirable characteristic of predictive maintenance algorithms of renewable energy systems. There are already some interesting solutions for the interpretation of ANNs models \cite{ahmad2020developing}. Thus, it is an interesting aspect of the domain to be further explored.  
    \item \textbf{Cloud Computing/Deployment}: Generally, the SCADA systems including the monitoring, control, and prognostic maintenance systems are deployed on-site. However, due the geo-graphic challenges involved in deploying such system on-site, which are generally located in the middle of ocean or on top of the mountains, and the advancement in cloud computing, the trend is shifting towards cloud deployment of such systems \cite{cloud_deployments}. There are already some efforts in this direction \cite{kande2017rotating,xiao2018prognostics}. 
    \item \textbf{Adversarially Robust Models}: Similar to other domains, ML/DL algorithms deployed in smart energy systems are also prone to several adversarial and security attacks especially in cloud deployments \cite{bor2019adversarial,sayghe2020evasion}. These attacks are intended to disturb the prediction capabilities of ML/DL algorithms. For instance, an adversarial attack on an ML/DL model for predictive maintenance may result in false alarms of faults in an energy system. The development of adversarially robust ML/DL models could be an interesting aspect of the predictive maintenance of renewable smart energy systems to be explored. 
    \item \textbf{Scarcity of Data}: Since, the occurrence of a fault in a power station is a rare event, the availability of faulty historic data for the training and testing of the model is a big challenge. The development of a publicly available dataset that includes both the clean and faulty SCADA data, specifically from the hydropower stations, could prove to be very fruitful and effective.
    \item \textbf{Novelty Detection}: The occurrence of faults in complex energy systems being a rare event has limited the availability of failure data. Consequently, it questions the very reliability and robustness of the ML/DL algorithms. This challenge could be addressed by using the technique of Novelty Detections. Hence, further research on the incorporation of novelty detection techniques in the prognostic and CBM systems needs to be carried out.
    \item \textbf{State-of-the-art Sensor Technologies}: The conventional CBM systems are mostly based on vibration and temperature analysis. Since most of these sensors/ accelerometers are mounted on the housings, they lead to a low signal-to-noise ratio, thereby affecting the efficiency of the prognostics and CBM systems. The use of miniature sensors close to the bearings could be an interesting development. Furthermore, the impacts of replacing orthodox sensing technologies with more novel techniques such as Barkhausen noise inspection and hydrogen sensors on the performance and efficiency of prognostics and CBM systems could be analyzed and exploited. 
    \item \textbf{Real-Time Prognostic Models}: Such prognostic models need to be developed that are capable of handling online data to get updated, and cater to the transient behaviors in the operational environment in order to increase their prediction accuracy, accordingly.
    
\end{itemize}

\begin{figure}[!h]
\centering
\graphicspath{ {./Figures/} }
\includegraphics[width=.89\linewidth]{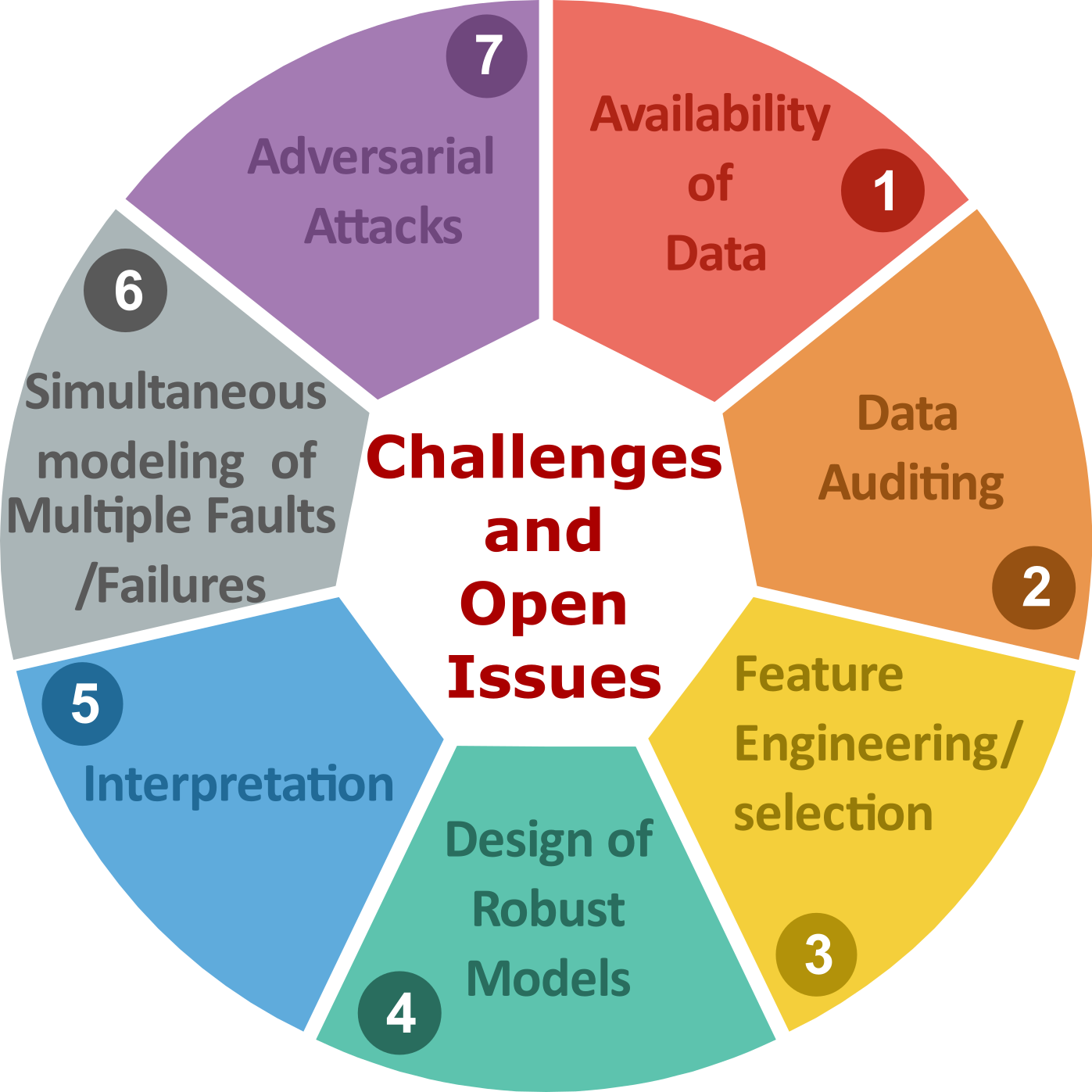}
\caption{A summary of key challenges and open issues in AI based prognostic maintenance of renewable energy systems.}
	\label{fig:challenges}
\end{figure}

Figure \ref{fig:challenges} summarizes the key challenges and open issues in the domain.

\section{Discussion and Lessons Learned}
\label{sec:discussion}
Prognosis is an important activity for predicting the RUL and optimization of the renewable energy system(s). Failures at such key installations can lead to extended downtimes thereby incurring economic losses and disrupting energy security. Moreover, the non-linear and non-stationary nature of these systems has induced a higher degree of uncertainty and complexity. Therefore, the selection of the most suitable and appropriate algorithm for the prognostics of these complex systems is very crucial. 

The existence of non-linear and non-stationary systems in the real world has induced a degree of uncertainty. Consequently, the selection of an appropriate prognostic method has become a very complex process. An optimum algorithm that takes into account all the uncertainties is very critical for the development of a robust prognostic system for real-life applications. Furthermore, generalization of the prognostic systems is a very difficult task because the condition data vary enormously, even for the systems having the same physical configuration but different operating environments and locations. For instance, the SCADA data for an on-shore wind turbine having the same specifications and configurations will vary from that of an off-shore wind turbine. Likewise, the SCADA data for hydro turbines operating in different vicinities will differ from each other. Also, the cases where the equipment has run to failure are very rare, hence the acquisition of real-life faulty data for training the algorithm is very difficult. Hence, future researches can be focused on developing a prognostic system that can efficiently model the uncertain and dynamic behavior of the equipment and are trained on real-life data acquired from the systems that are already in operation.  This will make the prognostic system more adaptable and will also address the conundrum of non-linearity and non-stationarity.

Some key lessons learned from this study are summarized as follows. 

\begin{itemize}
    \item The case studies published in the academic literature that illustrate the application of prognostic models in realistic operating environments having systems with overlapping failure modes are available, however, more work is required to be published in this area to verify the usefulness of prognostic models.
    \item The prognostics and RUL models have been classified into (i) Data-driven approaches,  (ii) Model-based approaches, and (iii) hybrid Approaches.
    \item There are several challenges associated with predictive maintenance of renewable energy systems, such as availability and quality of data, explainability, and adversarial attacks. 
    \item Availability of data and the level of accuracy required governs the selection of the appropriate models.
    \item Those scenarios where speed is more important than accuracy, data-driven models are preferred by the research community. 
    \item The selection of the modeling approach must complement the purpose. Since all the models are being developed based on some underlying assumptions and implementation constraints, it restricts their applicability to certain specific types of problems.
    \item All the models require data for defining the parameters, designing, and validation processes. However, the extent of data required varies from one model to another.
    \item Majority of the initial efforts rely on classical ML algorithms, which are trained on hand-crafted features. However, recently the trend shifted towards DL techniques. 
    \item These predictive maintenance algorithms are prone to adversarial and other security attacks. Thus, adversarially robust models need to be developed. 
    \item The data requirement for diagnostic modeling is different from that of prognostic modeling.
    \item All the models do not provide the confidence interval to their predictions. Most of the ANNs-based models cannot determine the confidence intervals for their estimates which on the other hand is required to manage the uncertainty and the associated risk.
    \item Different models have different levels of accuracy and precision to predict the RUL.
    \item The ability of a prognostic health management system pertinent to the prognosis of diversified faults and provide an estimation of the RUL in complex systems is still at the early stages of research and is likely to continue as an ongoing challenge.
    \item The prognostic models for the PV systems are mostly designed to diagnose and detect failures that are catastrophic and they do not monitor the system degradation over time or measure the RUL. Future researches can therefore be targeted towards the prognosis and health monitoring of the PV systems and measuring the degradation and RUL of the PV modules. 
\end{itemize}

\section{Conclusions}
\label{conclusion}
This paper reviewed the state-of-the-art models, techniques, and algorithms for prognostics and condition-based monitoring of electro-mechanical and renewable energy systems. The methodologies, advantages, and shortcomings of these models are briefly discussed. A detailed review of the latest researches conducted in the area of CBM and Prognostic Health Management of renewable energy systems including hydro, wind, and solar power projects is carried out. We also discussed some publicly available datasets for training and evaluation of predictive maintenance frameworks. Moreover, key challenges and future research directions are also identified.
The case studies published in the academic literature that illustrate the application of prognostic models in realistic operating environments having systems with overlapping failure modes are available, however, more work is required to be carried out in this area to verify the usefulness of prognostic models.

\bibliographystyle{unsrt}
\bibliography{AIonEdge}

\end{document}